\documentclass[10pt,twocolumn,letterpaper]{article}

\usepackage{wacv}
\usepackage{times}
\usepackage{epsfig}
\usepackage{graphicx}
\usepackage{amsmath}
\usepackage{amssymb}
\usepackage{booktabs} 
\usepackage{multirow}
\usepackage{enumitem}
\usepackage{makecell}
\usepackage{hyperref}

\wacvfinalcopy 

\def\wacvPaperID{****} 

\ifwacvfinal\fi
\begin{document}

\title{Ordinal Regression using Noisy Pairwise Comparisons for Body Mass Index Range Estimation}

\author{Luisa~F.~Polan\'{i}a \qquad Dongning~Wang \qquad Glenn M. Fung \\
American Family Insurance, Strategic Data \& Analytics, Madison, WI\\
{\tt\small \{lpolania, dwang1, gfung\}@amfam.com}
}



\maketitle
\ifwacvfinal\thispagestyle{empty}\fi

\begin{abstract}
   Ordinal regression aims to classify instances into ordinal categories. In this paper, body mass index (BMI) category estimation from facial images is cast as an ordinal regression problem. In particular, noisy binary search algorithms based on pairwise comparisons are employed to exploit the ordinal relationship among BMI categories. Comparisons are performed with Siamese architectures, one of which uses the Bradley-Terry model probabilities as target. The Bradley-Terry model is an approach to describe probabilities of the possible outcomes when elements of a set are repeatedly compared with one another in pairs. Experimental results show that our approach outperforms classification and regression-based methods at estimating BMI categories.
\end{abstract}

\section{Introduction}

Body mass index is a biometric that provides information about health condition and is frequently employed as a measure to diagnose obesity~\cite{Rome08}. It is defined as
\begin{equation}\label{eq6.4}
\text{BMI}=\frac{\text{Weight (kg)}}{(\text{Height (m)})^2}.
\end{equation}
The traditional way to measure BMI requires the presence of the subject and external elements, such as a measurement tape and a weight scale. Therefore, BMI measurement from facial images is of interest for many applications where there is no access to monitored measurement devices. For example, health-related analysis using profile images from social media ~\cite{Webe16, Koca17}, telemedicine kiosks to remotely diagnose patients~\cite{Lowe10} and face recognition~\cite{Wen14}. In the life insurance industry, BMI range estimation using face images has the potential to accelerate the underwriting process by alleviating the need for a medical exam. However, BMI estimation from face images is a challenging problem due to BMI distribution variations across races and ages~\cite{Koca17}, and due to the lack of information since a body-dependent measure is attempted to be estimated with only facial data.

Previous methods have been proposed for BMI estimation using facial images~\cite{Wen13, Rica06, Lee12, Koca17}. In~\cite{Wen13}, active shape models were employed for extraction of geometric features and three regression methods were used for the BMI estimation. Similarly, in~\cite{Lee12}, facial fiducial points were calculated for feature extraction, using a small dataset of 1124 face images. Convolutional neural networks (CNNs) have also been used for BMI estimation. In~\cite{Koca17}, the VGG-Net model and the VGG-face model were used for feature extraction.

In this paper, the problem of BMI category estimation is addressed. The same BMI categorization proposed by the World Health Organization~\cite{Worl06}, as indicated in Table \ref{table_BMI}, is used. A multi-class classification approach is not recommended for this problem, since it assumes independence between the class labels, which is not true for BMI categories since they have a strong ordinal relationship. Instead, the BMI category estimation problem is cast as an ordinal regression problem and addressed using a Noisy Binary Search (NBS)~\cite{Karp07} approach, where the goal is to insert the BMI associated to the test image into its proper place within the ordered sequence $S$, defined by the boundaries of the BMI categories. Even though it was suggested in~\cite{Karp07} that NBS algorithms could be employed for ranking problems, to the best of our knowledge, this is the first work that uses NBS for an ordinal regression application.

Noisy binary search relies on pairwise comparisons. Two deep learning-based models are used to make the comparisons. Both models consist of Siamese-type architectures~\cite{Chop05}. The differences between the architectures are related to the targets and the loss functions. One network uses binary classes as targets and the cross-entropy loss function. The other network uses the Bradley-Terry model probabilities~\cite{Brad52} as target and the Euclidean loss function.


The contributions of this paper are as follows:

- The application of NBS algorithms, which outperform classification methods based on CNNs and handcrafted features, to the problem of BMI category estimation.

- Two Siamese-type architectures to calculate pairwise comparisons. The architectures include modifications with respect to the traditional Siamese architecture~\cite{Koca17} used to learn similarity metrics. For example, both architectures incorporate the dot product between image feature vectors to further exploit correlation between the inputs. In addition, one of the architectures uses the Bradley-Terry model probabilities~\cite{Brad52} as target.

- To the best of our knowledge, this is the first work that addresses the BMI category estimation problem as an ordinal regression problem.

- This work uses the inmate active population dataset from the Florida Department of Corrections, which is the largest dataset that has ever been used for the problem of BMI estimation.


\section{Background}

This section describes the NBS problem and algorithms, the Siamese architecture typically used to learn similarity metrics, and the Bradley-Terry model.

\subsection{Noisy Binary Search}
\label{NBS_sec}
The goal of NBS is the same goal of the traditional binary search problem of inserting an element $x$ into its proper place within an ordered sequence $S=\{s_0,s_1, \ldots, s_{n-1}\}$ by comparing it with elements of the sequence. However, comparisons are noisy in NBS, \textit{i.e.}, gives the wrong result with a small probability. Therefore, each element $s_i$ of the sequence has an associated $p_i$ corresponding to the probability that $s_i<=x$. The empirical probability $\hat{p}_i$, a proxy for $p_i$, is estimated by performing multiple comparisons between $x$ and $s_i$.

The work in~\cite{Karp07} illustrates the NBS problem with the coin flip model in which each element from the sequence $S$ is assigned a coin whose heads probability, $p_i$, is unknown. However, the head probabilities are assumed to be ordered $p_0 \geq p_1\geq \ldots \geq p_{n-1}$ and we are allowed to toss the given coin to be able to estimate the empirical heads probability. If a coin is tossed $1/\epsilon^2$ times, then the probability that the estimated heads probability differs from $p_i$ by more than $\epsilon$ is bounded by a constant from Chernoff bound~\cite{Karp07}. The problem is solved when a pair of consecutive coins, $i$ and $i+1$, such that the interval $[\hat{p}_i, \hat{p}_{i+1}]$ contains the number $1/2$, is found. In this paper, two algorithms to solve the NBS problem are considered, namely the Naive Noisy Binary Search (NNBS) and the Interval Noisy Binary Search (INBS) algorithms.

\begin{table}[t]
\footnotesize
\renewcommand{\arraystretch}{1.3}
\caption{BMI categorization by the World Health Organization~\cite{Worl06}}
\label{table_BMI}
\centering
\begin{center}
\begin{tabular}{ |c|c| }
 \hline
 Category & BMI range (kg$/$m$^2$)\\  \hline
 Underweight & 16-18.5 \\
 Normal & 18.5-25 \\
 Overweight & 25-30 \\
 Moderately obese & 30-35 \\
 Severely obese & 35-40 \\
 \hline
\end{tabular}
\end{center}
\end{table}

\subsubsection{Naive Noisy Binary Search}
The NNBS algorithm is recursive and resembles the traditional binary search algorithm. It maintains a set of indexes $a$ and $b$ which are initialized to $s_0$ and $s_{n-1}$, respectively. At each iteration, it tests the sequence element midway between $a$ and $b$, denoted as $c$. If the calculated empirical probability associated to $c$, denoted as $\hat{p}_c$, is within $[1/2-\epsilon, 1/2+\epsilon]$, then the algorithm returns $c$. Otherwise, if $\hat{p}_c>1/2+\epsilon$, then index $a$ is updated with $c$ and $b$ remains unchanged. Similarly, if $\hat{p}_c<1/2+\epsilon$, then index $b$ is updated with $c$ and $a$ remains unchanged.  The process repeats until either $\hat{p}_c \in [1/2-\epsilon, 1/2+\epsilon]$, in which case $c$ is returned, or until $a$ equals $b$, in which case the algorithm returns $a$. Details of the NNBS algorithm can be found in~\cite{Karp07}.

\subsubsection{Interval Noisy Binary Search}
Interval Noisy Binary Search modifies the NNBS algorithm by allowing backtracking~\cite{Fala17}. It first builds a binary search tree of intervals such that the root node corresponds to sequence $S$. Each non-leaf node interval $I$ has two children corresponding to the left and right halves of $I$. The leaves of the tree are the intervals between consecutive sequence elements. The algorithm starts at the root of the binary search tree and at every non-leaf node corresponding to interval $I$, it checks if the element to be searched, $x$, belongs to $I$ by calculating the empirical probabilities associated to the sequence elements that define the boundaries of $I$. If either the empirical probability of the left boundary is  smaller than 0.5 or the empirical probability of the right boundary is greater than 0.5, the algorithm backtracks to the current node's parent. Otherwise, if 0.5 lies within the empirical probability of the boundaries, the algorithm checks if $x$ belongs to the left or right child by calculating the empirical probability associated with the  middle element of $I$. If it is greater than 0.5, then it moves to the right child, otherwise, it moves to the left child. At a leaf node, the algorithm checks if $x$ belongs to the corresponding leaf interval by maintaining a counter. The counter increases by one if 0.5 lies within the probability of the leaf interval boundaries. Otherwise, the counter decreases by one. If the counter becomes less than 0, the algorithm backtracks to the leaf's parent. The algorithm stops when the counter reaches a threshold $K_1$ and INBS returns the corresponding leaf node.

By following the above procedure, the algorithm may end up moving in a loop. If that is the case, the algorithm is run for a maximum of $K_2$ steps, saves all the visited sequence elements in a set $Q$, and runs NNBS on the set $Q$. Details of INBS can be found in~\cite{Fala17}.

\subsection{Siamese architecture}

Siamese convolutional networks have been widely employed to measure similarity in different applications, such as matching of image patches~\cite{Zago15}, face recognition~\cite{Chop05} and signature verification~\cite{Sign17}. All these problems fall into the category of matching problems. In ~\cite{LiuWB17}, Siamese networks were used to rank images in terms of image quality.

There are twin branches in a Siamese network that share the same architecture and the same set of weights. Let $z_i$ and $z_j$ denote the feature representations of the inputs provided by the last layers of the twin branches. Siamese architectures are typically trained with the contrastive loss function as follows
\begin{eqnarray}\label{eq6.18}
L(\theta)&=&\sum_{(z_i,  z_j) \in D} y_{ij}d_{ij}^2+(1-y_{ij})\text{max}(0, m^2-d_{ij}^2)  \nonumber\\
d_{ij}&=&\|z_i-z_j\|_2,
\end{eqnarray}
where $D$ is the set of feature representation pairs produced by the last layers of the Siamese network across all the inputs, $\theta$ are the weights of the network, $y_{ij} \in \{0,1\}$  is the label with 1 and 0 denoting a matching and a non-matching pair, respectively. The first term of the loss function penalizes matching pairs whose feature representations are far apart while the second term penalizes non-matching pairs whose feature representations are closer than a margin $m$.

\subsection{Bradley-Terry Model}
The Bradley-Terry model is a probability model used to predict the outcome of a comparison~\cite{Brad52}. Given a pair of individuals $i$ and $j$ drawn from some population, it estimates the probability that $i$ beats $j$ as $P(i \text{ beats } j)= {\gamma_{i}}/{(\gamma_{i}+\gamma_{j})}$,
where $\gamma_{i}$ and $\gamma_{j}$ are positive real-valued scores associated to individual $i$ and $j$, respectively. For example, $P(i \text{ beats } j)$ may denote the probability that player $i$  will win a game against player $j$ and $\gamma_{i}$ and $\gamma_{j}$ may represent player strengths or abilities.

\section{Method}
This section describes the proposed method to address the problem of BMI category estimation. Given an image $x$, the goal is to determine in which category from Table~\ref{table_BMI}, the BMI associated to image $x$ belongs to. We propose to address this problem with an NBS approach.

\subsection{Noisy binary search for BMI ordinal regression}
Let $S=\{16, 18.5, 25, 30, 35, 40\}$ be the sequence formed by the boundaries of the BMI categories. Each element $s_i$ of $S$ is represented by a pool of images, referred to as anchors, such that their BMI falls in the range $[s_i-\gamma, s_i+\gamma]$, where $\gamma$ is a small constant, and therefore, the pool of images have BMIs approximately equal to $s_i$. The goal is to insert the BMI of $x$, denoted as $x_{BMI}$ into its proper place within $S$, by performing comparisons between $x$ and the anchors in an NBS fashion. For this purpose, a comparison operator is built such that it outputs 1 if it predicts that the anchor image has a BMI smaller or equal than $x_{BMI}$ and 0 otherwise. The comparison operator is noisy, \textit{i.e.} it gives the wrong result with a small probability.

The proposed approach will be explained by using an analogy with the coin flip model presented in Section \ref{NBS_sec}. The equivalent of flipping the coin assigned to $s_i$ is to randomly select an image from the anchors associated to $s_i$ and run it with $x$ through the comparison operator. The output of the operator, either 1 or 0, is the equivalent of the flip result, either head or tail. In Section \ref{NBS_sec}, it was described that the coin assigned to $s_i$ was flipped multiple times to calculate the empirical probability $\hat{p}_i$. Similarly, the comparison operator is run with several randomly selected anchor images assigned to $s_i$, to calculate the empirical probability $\hat{p}_i$ that $s_i \le x_{BMI}$. Those probabilities are used by the NBS algorithms to predict the right place for $x_{BMI}$ within the order sequence $S$, as explained in Section \ref{NBS_sec}.

\subsection{Comparison operator}
\label{comparison_operator}
This section presents comparison operators that are built with Siamese-type networks. The twin branches are truncated versions of traditional CNN architectures. Specifically, the AgeNet~\cite{Levi15} and the VGG architecture, which was used in~\cite{Koca17} for BMI regression, are employed.

AgeNet consists of 3 convolutional layers, three fully connected (FC) layers and a softmax layer. The first, second, and third convolutional layers contain 96 filters of size $3 \times 7 \times 7$, 256 filters of size $96 \times 5 \times 5$, and 384 filters of size $256 \times 3 \times 3$, respectively. The first two FC layers contain 512 neurons each and the last layer contains 8 neurons. Truncated versions of the AgeNet architecture, built by excluding the softmax and the last FC layer, are used as the twin branches of the Siamese network. The motivation for using a small and simple architecture, such as AgeNet, is that learning to compare image pairs to predict which image has higher BMI is intuitively easier than learning the nominal BMI category.

The VGG architecture, which has more representational power than AgeNet, is also employed, at the expense of increasing the memory and computational cost. VGG contains 16 layers, 13 convolutional layers and 3 FC layers followed by a softmax. All the convolutional layers have a receptive field of size $3\times3$ and are followed by a ReLU layer. A stack of 13 convolutional layers are followed by three FC layers, where the first two have $4096$ channels each, and the third has a number of channels that depends on the classification task. Another configuration for the twin branches of the Siamese network used in this paper is a truncated version of the VGG architecture, built by excluding the softmax and 2 last FC layers.

\begin{figure}[t]
\centering{ 
\includegraphics[width = 0.85\columnwidth]{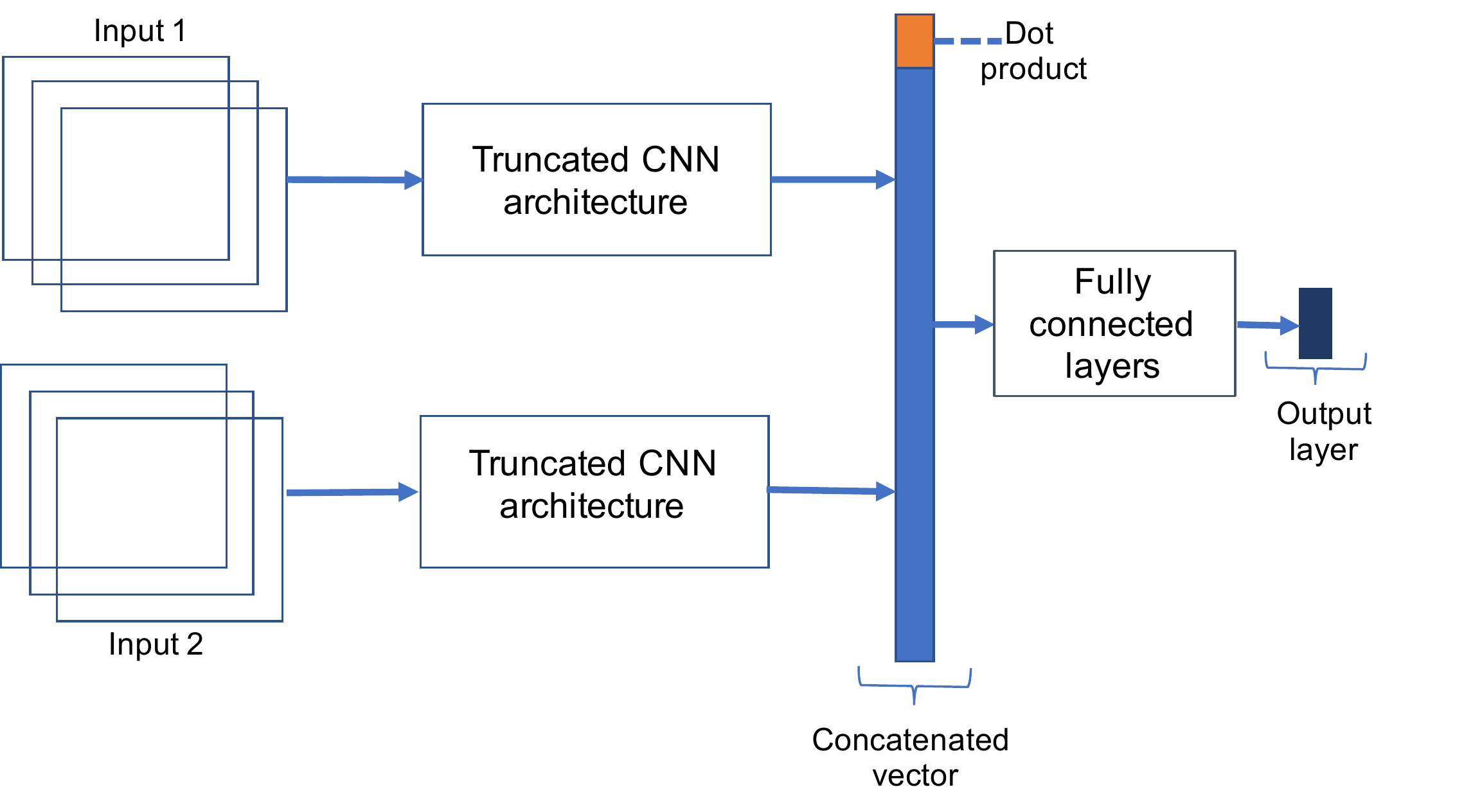}}
\caption{Schematic of the Siamese architecture} \label{architecture}
\end{figure}

Feature outputs from the twin branches are concatenated. To further exploit the correlation of the features, the dot product between the features is included in the concatenation vector. The dot product is a modification with respect to the traditional Siamese architecture~\cite{Koca17}. In the case of the AgeNet-based Siamese network, two FC layers follow the concatenation. The first and second FC layers contain 512 and 1 neurons, respectively.  In the case of the VGG-based Siamese network, three FC layers follow the concatenation. The first and second FC layers contain 2048 and 1024 neurons, respectively, and are followed by ReLU and dropout layers each. For both networks, the last FC layer contains a single output which is fed to a sigmoid function. The architecture of the Siamese network is illustrated in Fig. \ref{architecture}.

The comparison operator can be represented with the function $f(a_j^{s_i}, x; \theta)$, where $\theta$ denotes the network parameters and $a_j^{s_i}$ and $x$ denote the inputs, which are the anchor image used for the $j$th comparison and the given image, respectively. The function takes the value 1 when the network predicts that the BMI of $a_j^{s_i}$ is smaller or equal than the BMI of $x$. Otherwise, it outputs 0. Let $h_i$ denote the number of comparisons used to calculate the empirical probability $\hat{p}_i$ that $s_i\le x_{BMI}$, then the empirical probabilities that are fed to the NBS algorithms are defined by

\begin{equation}\label{eq6.4}
\hat{p}_i=\frac{ \sum_{j=0}^{h_i-1} f(a_j^{s_i}, x; \theta)}{h_i} \text{,  } i=0,\ldots, n-1.
\end{equation}

\subsubsection{Training modes}

Two training modes for the Siamese network are used in this paper. Mode I uses the binary classes 1 and 0 as targets, where class 1 means that the BMI of the anchor input image is smaller or equal than $x_{BMI}$ and class 0 means otherwise. Mode I uses the cross-entropy loss function, which is defined as
\begin{equation}\label{eq6.5}
L(\theta)=-\frac{1}{N_{t}}\sum_i^{N_{t}}[q_i\text{log}{g}_i+(1-q_i)\text{log}(1-{g}_i)],
\end{equation}
where $N_t$ is the number of training image pairs, $q_i$ is the truth label for the $i$th image pair and ${g}_i$ is the corresponding predicted probability. Mode II uses the Bradley-Terry model probabilities as target and the Euclidean loss function. The Bradley-Terry model is frequently used to model the outcome of games~\cite{Brad52} and the motivation for using it in this paper is that comparison between a given image and the anchors associated to a sequence element have some similarity with the outcome of a game in the sense that the comparison operator predicts which subject from the input images wins at having higher BMI. The randomness in the outcome of a game comes from the fact that the same player may perform differently at different times, while the randomness of the comparison operator at predicting if $s_i\le x_{BMI}$ comes from the fact that an anchor image is randomly selected from the pool of anchors at each comparison. The equivalent of the ability score $\gamma_i$ associated to player $i$ is the BMI associated to the image. Therefore, the Bradley-Terry model probabilities, adapted to our problem, are defined by

\begin{equation}\label{eq6.4}
P(x \text{ beats } a_j^{s_i})=\frac{x_{BMI}}{x_{BMI}+s_i}, \text{  } i=0,\small{\ldots}, n-1, \text{  } j=0, \small{\ldots}, h_i-1,
\end{equation}
where, as before, $a_j^{s_i}$ denotes the randomly selected image from the pool of anchor images associated to $s_i$ to perform the $j$th comparison.

The output probabilities of the mode II-trained network are mapped to binary outputs by using the criteria that if the probability is greater or equal than 0.5, then $f(a_j^{s_i}, x; \theta)=1$. Otherwise, $f(a_j^{s_i}, x; \theta)=0$.

\subsubsection{Training of the Siamese networks}
\label{training}
The procedure to build the training pairs for the networks is as follows. The entire dataset is first divided into training, denoted as training dataset I, validation and testing datasets. Images whose BMI is within the range $[s_i-\gamma, s_i+\gamma]$ for each sequence element $s_i$ are extracted from the training dataset I to build the anchor dataset. Let training dataset II denote the remaining set of training dataset I after extracting the anchors. A training budget $b_i$ is assigned to each $s_i$ and represents the number of training pairs assigned to $s_i$. A training pair is built by randomly selecting an image from the training dataset II and an anchor image. The budget $b_i$ is equally distributed among the anchors belonging to $s_i$. The set of training pairs form the dataset used to train the Siamese networks. The same procedure is followed to build the validation pairs, but starting from the validation dataset.

For training the Siamese architectures, faces are first detected using the algorithm in~\cite{Math14} and cropped to the size $224\times 224$. The top branches of the AgeNet-based and VGG-based Siamese networks are initialized with the weights of the original AgeNet~\cite{Levi15} and VGG-Face models~\cite{Park15}, respectively. The FC layers following the feature concatenation are initialized with the Xavier method~\cite{Gloro10}. For the AgeNet-based network, the Adam optimizer~\cite{King14} with a base learning rate of $1\times10^{-4}$ and with default momentum values $\beta_1=0.9$ and $\beta_2=0.999$ is used for training with 300 samples per mini-batch. For the VGG-based network, the weights of the first 10 convolutional layers are kept frozen during training. Also, the Adam optimizer with a base learning rate of $1\times10^{-5}$ and with default momentum values $\beta_1=0.9$ and $\beta_2=0.999$ is used with 30 samples per mini-batch only due to memory constraints. The dropout probability is set to 0.5. Training stops when the loss on the validation pairs stops decreasing.

\section{Experimental Results}
In this section, we present details of the database, ordinal regression quality metrics, and experiments used to evaluate the proposed method (code is available at \url{https://github.com/lfpolani/QI_ordinal_regression}).

\subsection{Database}
The Florida Department of Corrections is a downloadable database featuring records for all the inmates currently incarcerated in the Florida state prison system~\cite{Flori}. The database contains weight and height information. The dataset is filtered to consider subjects with BMI in the range 16-40, which corresponds to the categories of interest. Samples outside this range are rare. For the experiments, 67756, 9045, and 20000 samples  are randomly selected for training, validation and testing, respectively.

\begin{figure}[t]
\centering{ 
\includegraphics[width = .9\columnwidth]{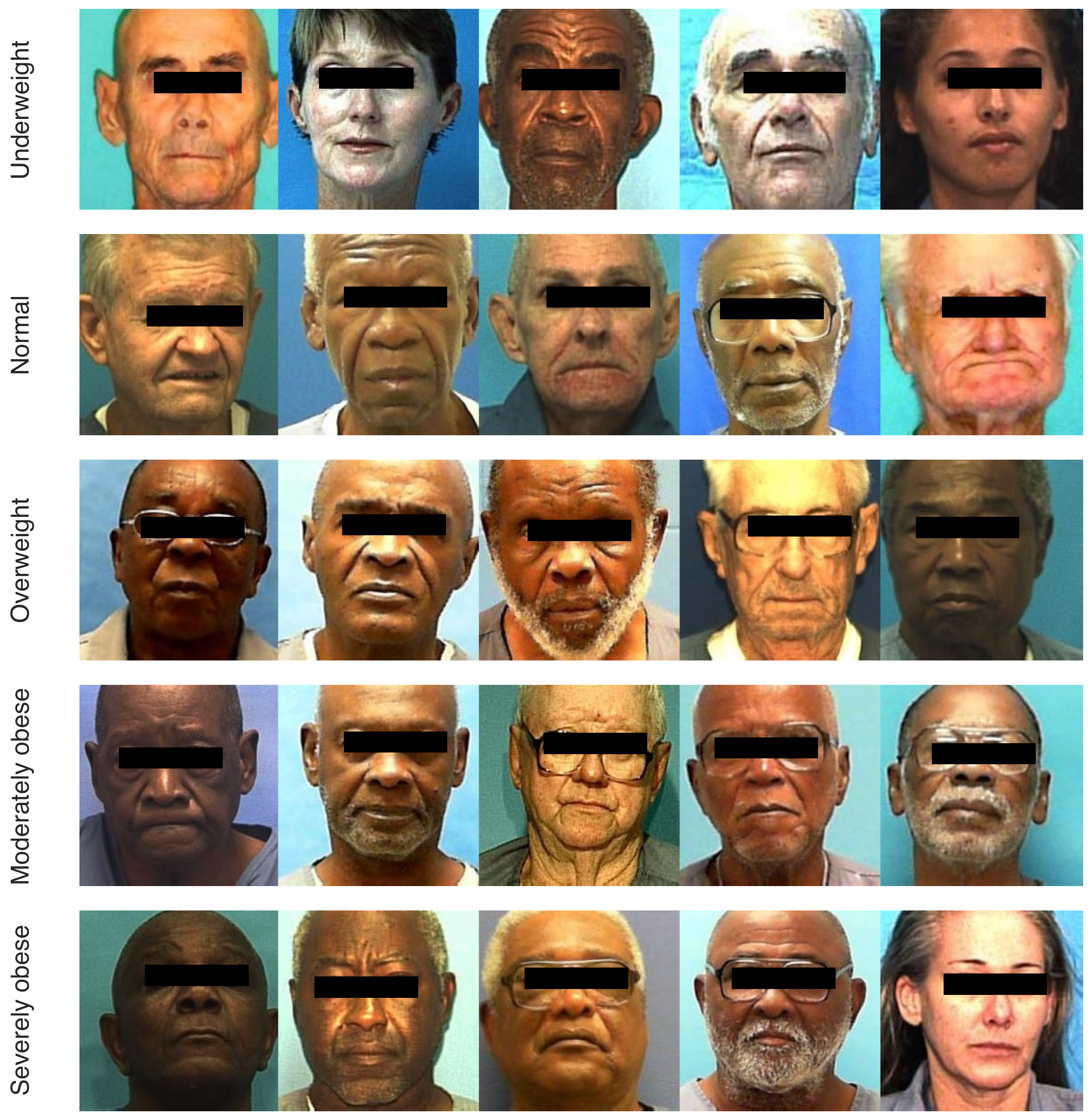}}
\caption{Image samples for each BMI category} \label{pictures}
\end{figure}

Fig. \ref{pictures} shows randomly selected images for each BMI category. Note that distinguishing images from consecutive BMI categories is visually challenging; especially distinguishing between normal and overweight images, which correspond to nearly 75\% of the entire dataset.

\subsection{Performance metrics}
Three metrics commonly used for ordinal regression problems \cite{Sanc13} are used to evaluate the performance of the proposed method, namely accuracy (ACC), Mean Absolute Error (MAE), and the Kendall rank correlation coefficient ($\tau$). The accuracy is the fraction of correctly classified samples. Let $0,1, \ldots, 4$ be the labels of the underweight, normal, overweight, moderately obese, and severely obese BMI categories, respectively. Let $y_i$ and $\tilde{y}_i$  denote the ground truth and predicted label for the testing sample $x_i$, respectively, and $m$ denote the total number of samples in the testing set. Then, the MAE is defined by $\frac{1}{m}\sum_{i=0}^{m-1}|\tilde{y}_i-y_i|$. The Kendall rank correlation coefficient is used to measure the correlation between the rankings of the ground truth and predicted labels. It is defined as
\begin{equation}\label{AMAE}
\tau=\frac{{\sum_{0\leq i<j\leq m-1}^{m-1}} {C((\tilde{y}_i, y_i),(\tilde{y}_j, y_j))}}{m(m-1)/2},
\end{equation}
where $C(\cdot)$ is the concordance indicator function, defined by
\begin{equation}\label{concordance}
C((\tilde{y}_i, y_i),(\tilde{y}_j, y_j)) = \left\{
        \begin{array}{ll}
            -1 & \text{if} (\tilde{y}_i-\tilde{y}_j)({y}_i-{y}_j)<0\\
            1 & \text{if} (\tilde{y}_i-\tilde{y}_j)({y}_i-{y}_j)>0.
        \end{array}
    \right.
\end{equation}
The Kendall rank correlation coefficient ranges from -1 to 1. For the perfect agreement and disagreement between the rankings of the ground truth and predicted labels, $\tau$ takes values 1 and -1, respectively. For completely independent rankings, $\tau$ has value 0.


For evaluating the performance of the Siamese networks, the accuracy and the area under the curve (AUC) are employed.

\subsection{Performance of the Siamese architecture}

This section evaluates the performance of the AgeNet-based Siamese network trained using the validation pairs described in Section \ref{training}. The training budget per sequence element $b_i$ is set to 150000. The value of $\gamma$ needed to build the anchor set is set to 0.3. The classification threshold for the mode I-trained network is chosen such that the difference between the true positive rate and the false positive rate is maximized. This criteria leads to a classification threshold of 0.548. The classification threshold of the mode II-trained network is set to 0.5 to take advantage of the symmetry of the Bradley-Terry model around 0.5.


Table \ref{siamese} shows the performance of the networks per sequence element. Note that sequence elements 16 and 40 are not included because the BMI of the dataset always lies between those two numbers due to pre-filtering.  The mode II-trained network outperforms the mode I-trained network for all the different sequence elements, except for 25. For mode I, the ground truth of the validation pairs associated to $s_i=18.5$ are mostly 1 since most subjects have a BMI greater than 18.5, and therefore, it is expected that the ACC associated to 18.5 should be high. Similarly, the ACC for $s_i=35$ is expected to also be high since the validation pairs are highly unbalanced for that case as well. The AUC is known to be a better performance metric for unbalanced datasets~\cite{Bhow12}. Table \ref{siamese} shows that the AUC attained by the Siamese networks mostly ranges from 0.7 to 0.8.





\begin{table}
\centering
\caption{Performance of the AgeNet-based Siamese network}
\label{siamese}
  \begin{tabular}{|c|c|c|c|c|}
    \hline
    \multirow{2}{*}{Sequence element} &
      \multicolumn{2}{c|}{Mode I} &
      \multicolumn{2}{c|}{Mode II} \\
    \cline{2-5}
    & ACC & AUC & ACC & AUC \\
    \hline
    18.5& 0.997 & 0.692 & 0.997 & 0.798 \\
    25& 0.71 & 0.725 & 0.694 & 0.712 \\
    30& 0.748 & 0.729 & 0.762 & 0.745 \\
    35& 0.933 & 0.775 & 0.946 & 0.788 \\
    \hline
  \end{tabular}
\end{table}

\subsection{Performance of the Noisy Binary Search algorithms}
\label{NBS_per}
The performance of both NNBS and INBS is evaluated in this section using the testing dataset. First, the performance of the algorithms is analyzed as a function of the comparison budget. Then, the NBS algorithms are compared with the AgeNet and VGG classification networks and with a handcrafted feature-based method. For the NNBS algorithm, the value of $\epsilon$ is set 0.03. The part of the INBS algorithm that runs the NNBS on $Q$ also uses $\epsilon=0.03$. The parameters $K_1$ and $K_2$ are set to $3 \log n$ and $12\log n$, where $n$ is the number of sequence elements.

\begin{figure*}[t]
\centering{ 
\includegraphics[width = 0.98\textwidth]{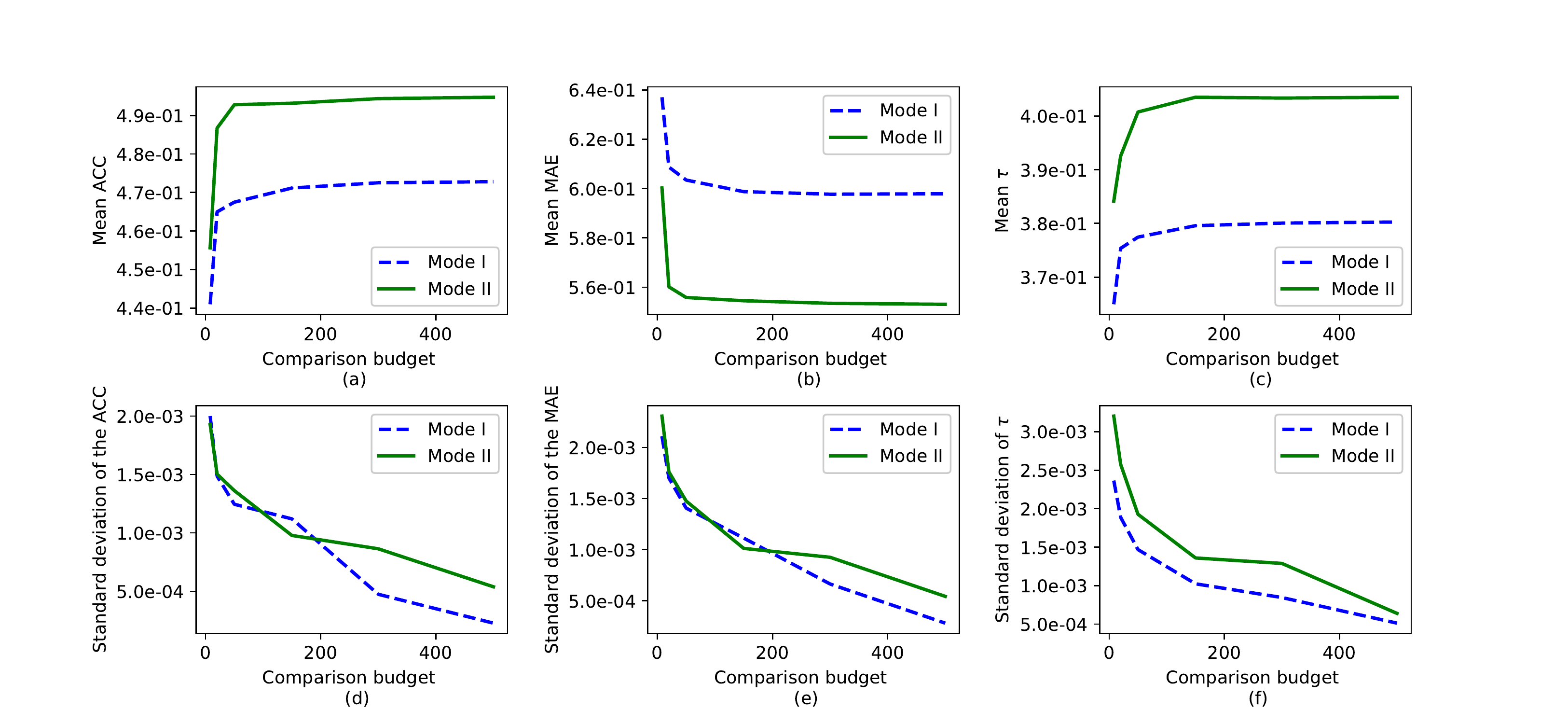}}
\caption{Performance of the NNBS algorithm as a function of the comparison budget $H$.} \label{NNBS_per}
\end{figure*}

\begin{figure*}[t]
\centering{ 
\includegraphics[width = 0.97\textwidth]{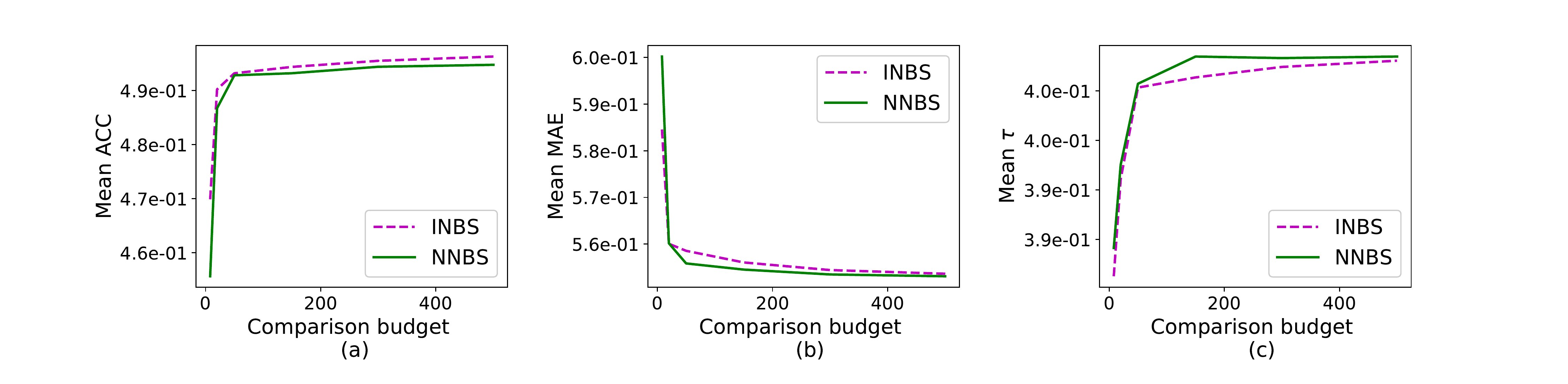}}
\caption{Comparison of the NNBS and INBS algorithms.} \label{INBS per}
\end{figure*}

\subsubsection{Performance evaluation as a function of the comparison budget}

A budget $H$ is assigned to the NBS algorithm to perform pairwise comparisons. The fraction of the budget $H$ assigned to a sequence element $s_i$, denoted as $h_i$, depends on the performance of the Siamese network at comparing with that particular sequence element. Thus, $h_i$ is defined as
\begin{equation}\label{budget}
h_i=\frac{1-\text{AUC}_i}{\sum_{k=1}^{n-2} (1-\text{AUC}_k)}, \hspace{.2cm} i=1, \ldots, n-2
\end{equation}
where $\text{AUC}_i$ is the AUC associated with the performance of the Siamese network at predicting if the BMI of a given image is greater or equal than $s_i$ (results shown in Table \ref{siamese}). Note that a budget is not assigned to $s_0$ and $s_{n-1}$ because the BMI of the testing samples always lies between $s_0$ and $s_{n-1}$ due to pre-filtering of the data.

Figure \ref{NNBS_per} illustrates the performance of the NNBS algorithm by varying the budget $H$ using the AgeNet-based Siamese network. The results in Figs. \ref{NNBS_per}(a-c) correspond to the mean of the metrics across repetitions using different random anchor sampling at each repetition. Similarly, Figs. \ref{NNBS_per}(d-f) correspond to the standard deviation of the metrics across repetitions. The number of repetitions is set to 50. As expected, Figs. \ref{NNBS_per}(a-c) show that the performance of the NNBS algorithm improves as the comparison budget increases. However, it improves very slowly for $H>50$. The mode II-trained network outperforms the mode I-trained network for the entire comparison budget range, which suggests that using the Bradley-Terry model probabilities is better than using binary labels as target. As the comparison budget increases, the variance of the predicted empirical probabilities across repetitions decreases, and therefore, the standard deviation of the performance metrics also decreases.

Figure \ref{INBS per} compares the performance of the NNBS and the INBS algorithms as a function of $H$. Note that in the case of the INBS algorithm, the definition of the budget fraction in \eqref{budget} applies at every step of the algorithm, but since backtracking may happen, the overall number of comparisons with $s_i$ may exceed $h_i$. A smaller comparison budget implies more variance in the calculated empirical probabilities across comparisons, and therefore, for small values of $H$, it is is expected that the INBS algorithm backtracks to the parent node more often to retry comparisons and improve performance than for large values of $H$. This behavior reflects in Fig. \ref{INBS per}(a), where the the INBS algorithm exhibits larger mean accuracy  and smaller mean MAE than the NNBS algorithm  for $H\le20$. Interval Noisy Binary Search also attains higher accuracy than NNBS for $H>20$, but the difference is marginal for that case since there is less variance in the calculated empirical probabilities, and therefore, less backtracking. The MAE metric attained by INBS is smaller than that of NNBS for $H=8$, but slightly bigger for $H>20$. Regarding the $\tau$ metric, it achieves almost the same values for both algorithms when $H<50$, but it is smaller for INBS than for NNBS when $H\ge50$. An explanation for the behavior of the MAE and $\tau$ metrics for $H\ge50$ is that for a relatively large number of comparisons, less backtracking is expected and if the counter does not reach the threshold $K_1$, the INBS reduces to NNBS applied to only the visited elements, instead of the entire sequence $S$. From Fig. \ref{INBS per}, we conclude that choosing INBS over NNBS is only worthy when a low comparison budget per iteration is required.

\subsubsection{Increasing the representational power of the Siamese Network}

\begin{table}
\centering
\caption{Comparison between the AgeNet-based and VGG-based Siamese networks}
\label{AgeNet_VGG}
  \begin{tabular}{|c|c|c|c|c|}
    \hline
    \multirow{2}{*}{Sequence element} &
      \multicolumn{2}{c|}{AgeNet-based} &
      \multicolumn{2}{c|}{VGG-based} \\
    \cline{2-5}
    & ACC & AUC & ACC & AUC \\
    \hline
    18.5& 0.997 & 0.692 & 0.997 & 0.743\\
    25& 0.71 & 0.725 & 0.7 & 0.739 \\
    30& 0.748 & 0.729 & 0.772 & 0.758\\
    35& 0.933 & 0.775 & 0.945 & 0.822  \\
    \hline
  \end{tabular}
\end{table}

\begin{figure*}[t]
\centering{ 
\includegraphics[width = \textwidth]{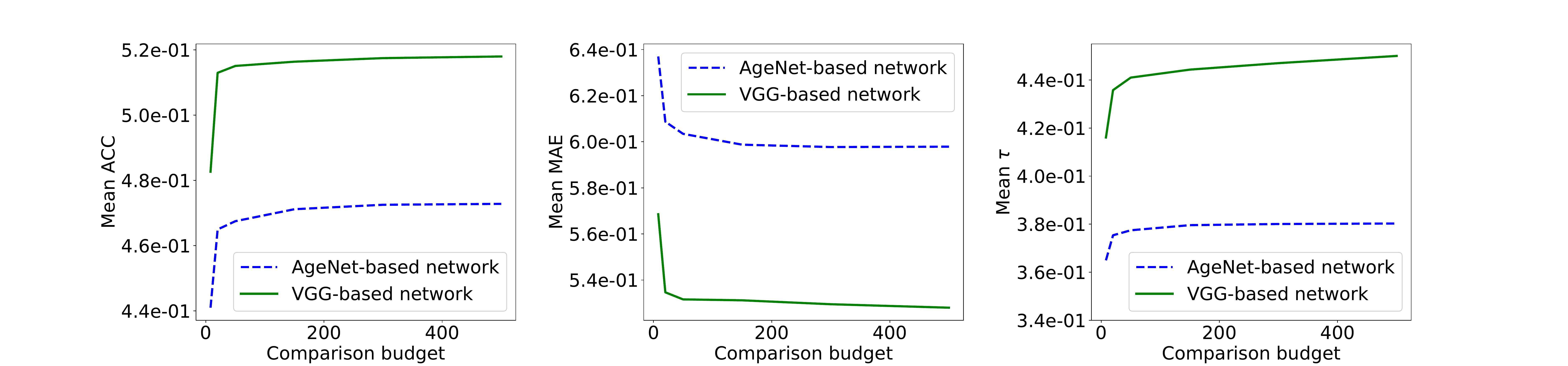}}
\caption{Performance of the NNBS algorithm using the AgeNet-based and VGG-based Siamese networks} \label{agenet_vgg_per}
\end{figure*}

An experiment to compare the performance of the NNBS algorithm using the AgeNet-based and VGG-based Siamese networks is presented in this section. The training mode I is used for the experiment. The INBS is not included in this experiment because it was shown in Section \ref{NBS_per} that it only outperforms NNBS when the comparison budget is low.

Table \ref{AgeNet_VGG} compares the performance of the AgeNet-based and VGG-based Siamese networks at predicting if the BMI of a given image is greater or equal than a sequence element, using the validation pairs described in Section \ref{training}. The results indicate that the VGG-based network outperforms the AgeNet-based network, which is expected given that VGG has more representational power than AgeNet. Only for the sequence element 25, the ACC obtained with the VGG-based network is slightly smaller than the ACC of the AgeNet-based network.

Fig. \ref{agenet_vgg_per} compares the performance of the NNBS algorithm using the AgeNet-based and VGG-based Siamese networks. As expected, using the VGG-based network in the NNBS algorithm leads to better results for the entire comparison budget range. For both networks, the performance improves slowly when the comparison budget exceeds 50.


\subsubsection{Performance comparison with classification-based and regression-based approaches}

In this section, the proposed method is compared with the AgeNet and VGG classification networks, with a handcrafted feature-based method, and with a VGG regression network followed by BMI category mapping. Comparison with previous results is not presented in this section because this is the first work that addresses BMI category estimation using the category definitions in Table \ref{table_BMI}.

The AgeNet and VGG classification networks employed in this experiment use the original AgeNet and VGG architectures presented in Section \ref{comparison_operator}, respectively, with the exception of the last FC layer which uses 5 outputs. Similarly, the VGG regression network uses the same original VGG architecture presented in Section \ref{comparison_operator}, with the exception of the last FC layer which uses 1 output and the softmax layer which is removed. The VGG regression network uses the  Euclidean loss function instead of the cross-entropy loss function. The training procedure for the networks is similar to that described in Section \ref{training}. Faces are cropped to the size of $224 \times 224$. The AgeNet and VGG networks are initialized with the weights of the original AgeNet model~\cite{Levi15} and VGG-Face model~\cite{Park15}, respectively, except for the last FC layer, which is initialized  with  the Xavier method~\cite{Gloro10}. The same optimizer, base learning rate, batch size, dropout factor and training stopping criteria of the AgeNet-based and VGG-based Siamese networks are employed for the AgeNet and VGG classification and regression networks, respectively. The output of the VGG regression network is mapped to the corresponding BMI category using the definitions in Table \ref{table_BMI}.

The handcrafted feature-based method uses the same geometric features proposed in~\cite{Wen13} for BMI regression. A multiclass linear-kernel SVM is employed for the BMI range estimation. The penalty parameter $C$ of the error term is estimated using grid search, which leads to $C=1$.

Table \ref{class_comp} compares the performance of the NNBS and INBS algorithms using a budget $H=500$ with the performance of the AgeNet and VGG classification networks, with the handcrafted feature-based method, and with the VGG regression network followed by BMI category mapping. The value of $H=500$ is selected for the comparison since it corresponds to the best performance achieved by the NBS algorithms. Note that the metrics in Table \ref{class_comp} for the NBS algorithms correspond to the mean values across repetitions using different random anchor sampling at each repetition. Results show that NBS algorithms outperform  classification-based methods by exploiting the ordinal nature of the BMI range estimation problem. The NBS algorithms using the AgeNet-based and VGG-based Siamese networks outperform the results of the AgeNet and VGG classification networks, respectively. Note that the $\tau$ metric results indicate that the rankings of the ground truth are much more correlated with the rankings of the labels predicted by the proposed methods than with the rankings of the labels predicted by the handcrafted feature-based method. The proposed methods also outperform the VGG regression network followed by BMI category mapping.

\begin{table}
\centering
\caption{Comparison of the NBS algorithms with classification-based methods}
\label{class_comp}
  \begin{tabular}{|c|c|c|c|}
    \hline

    Method & ACC & MAE & $\tau$  \\
    \hline
     \makecell{Handcrafted \\feature-based method}& 0.45 & 0.615 & 0.13 \\
    AgeNet (classification)& 0.467 & 0.616 & 0.362 \\

    NNBS (AgeNet, Mode I)& 0.473 & 0.598 & 0.38  \\
    NNBS (AgeNet, Mode II)& 0.495 & 0.553 & 0.403  \\
    INBS (AgeNet, Mode II)& 0.496 & 0.554 & 0.403  \\
    VGG (classification)& 0.494 & 0.555 & 0.45\\
\makecell{VGG (regression) + \\BMI category mapping} & 0.481 & 0.589 & 0.382\\
    NNBS (VGG, Mode I)& 0.518  & 0.528 & 0.45\\
    \hline
  \end{tabular}
\end{table}

\section{Conclusions}
Noisy binary search has been studied extensively in the area of computer science. In this paper, it was shown how the ability of NBS algorithms to exploit the ordinal nature of a problem can be leveraged to address the problem of BMI range estimation. As NBS relies on pairwise comparisons, two Siamese networks were proposed to perform the comparisons. The motivation for using a method based on pairwise comparisons was that predicting which subject from two images has higher BMI is intuitively
easier than learning the nominal BMI category.

{\small
\bibliographystyle{ieee}
\bibliography{egbib}
}

\end{document}